%% file: main.tex
\documentclass[journal,twoside,web]{ieeecolor}

\usepackage{generic}
\usepackage{cite}
\usepackage{amsmath,amssymb,amsfonts}
\usepackage{graphicx}
\usepackage{bmpsize}
\usepackage{textcomp}
\usepackage{subfigure}
\usepackage{epsfig} 

\usepackage{url} 
\usepackage{lcsys}
\usepackage{algorithm} 
\usepackage[noend]{algpseudocode} 
\usepackage{gensymb} 
\usepackage{svg} 
\usepackage{xcolor} 
\usepackage{siunitx} 
\usepackage{kotex}

\makeatletter
\let\NAT@parse\undefined
\makeatother
\usepackage[pagebackref,breaklinks,hidelinks]{hyperref}
\begin{document}

\def\BibTeX{{\rm B\kern-.05em{\sc i\kern-.025em b}\kern-.08em
    T\kern-.1667em\lower.7ex\hbox{E}\kern-.125emX}}
\markboth{\journalname, VOL. XX, NO. XX, XXXX 2017}
{Author \MakeLowercase{\textit{et al.}}: Preparation of Papers for IEEE Control Systems Letters (August 2022)}

\title{
Model Parameter Identification via a Hyperparameter Optimization Scheme\\ for Autonomous Racing Systems
}
\thispagestyle{empty} 

\author{
Hyunki Seong$^{1}$,
Chanyoung Chung$^{2*}$,
and David Hyunchul Shim$^{1}$
\thanks{This research was financially supported by the Institute of Civil Military Technology Cooperation funded by the Defense Acquisition Program Administration and Ministry of Trade, Industry and Energy of Korean government under grant No. UM22206RD2.}
\thanks{$^*$Corresponding author}
\thanks{$^{1}$ The authors are with the School of Electrical Engineering, Korea Advanced Institute of Science and Technology, Daejeon, South Korea. (email: hynkis@kaist.ac.kr;geninfty@kaist.ac.kr)}
\thanks{$^{2}$ Chanyoung Chung is with the JPL Science Division, NASA, California, USA. (email: chanyoung.chung@jpl.nasa.gov)}
}

\maketitle
\thispagestyle{empty}

\begin{abstract}
\input{sections/0_abstract}
\end{abstract}

\begin{IEEEkeywords}
Data-driven control, hyperparameter optimization, autonomous vehicle
\end{IEEEkeywords}

\input{sections/1_introduction.tex}
\input{sections/3.1_model_learning.tex}
\input{sections/3.2_vehicle_dynamics_model.tex}

\input{sections/3.3_model_based_vehicle_control.tex}
\input{sections/4.experiment.tex}

\input{sections/5.conclusion.tex}

\bibliographystyle{IEEEtran}
\bibliography{main}

\begin{figure}[t]
\centering
\includegraphics[width=0.48\textwidth]{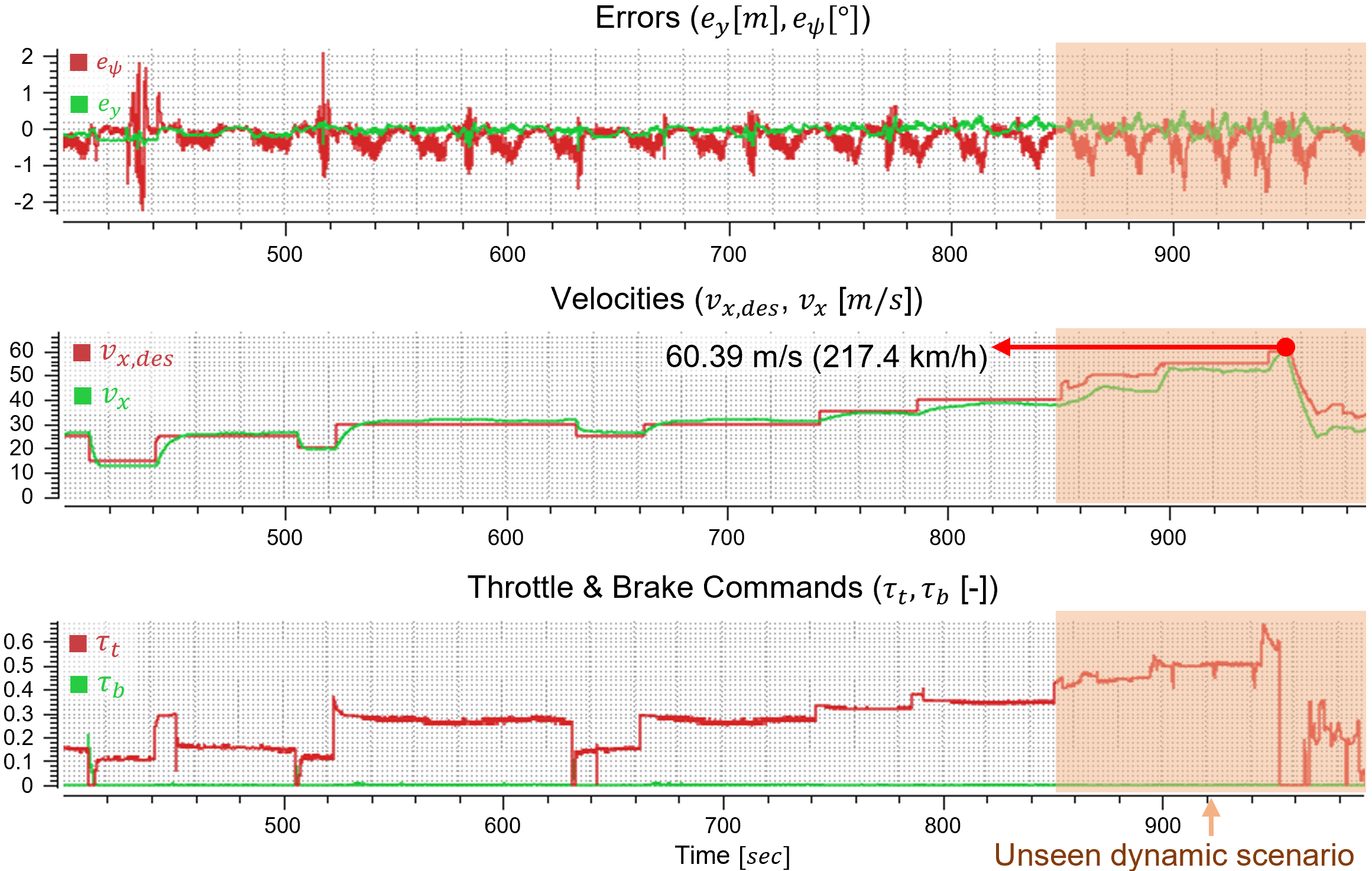}
\caption{
Results of the position/orientation errors, velocity control, and throttle/brake controls in LVMS.
}
\label{fig:result_LVMS_tracking_perform_only}
\vspace{-1.3em}
\end{figure}

\end{document}

%% file: sections/0_abstract.tex
In this letter, we propose a model parameter identification method via a hyperparameter optimization scheme (MI-HPO).
Our method adopts an efficient explore-exploit strategy to identify the parameters of dynamic models in a data-driven optimization manner. We utilize our method for model parameter identification of the AV-21, a full-scaled autonomous race vehicle.
We then incorporate the optimized parameters for the design of model-based planning and control systems of our platform.
In experiments, MI-HPO exhibits more than 13 times faster convergence than traditional parameter identification methods. Furthermore, the parametric models learned via MI-HPO demonstrate good fitness to the given datasets and show generalization ability in unseen dynamic scenarios.
We further conduct extensive field tests to validate our model-based system, demonstrating stable obstacle avoidance and high-speed driving up to \SI{217}{km/h} at the Indianapolis Motor Speedway and Las Vegas Motor Speedway.
The source code for our work and videos of the tests are available at \url{https://github.com/hynkis/MI-HPO}.

%% file: sections/1_introduction.tex
\section{Introduction}
\label{sec:introduction}
\IEEEPARstart{U}{nderstanding} the system model is essential for robotic applications, especially safety-critical autonomous systems. In particular, at high-speed driving, various dynamics elements such as chassis, tires, or engines become crucial to implement high-speed autonomy.
Model-based optimal control\cite{lewis2012optimal,jung2021game} is well-suited for handling those factors and is widely used to design dynamics system control.
By leveraging physics-based parametric dynamic models, it optimizes driving behavior with respect to a designed objective function and enables safe and reliable control system design.

Despite the success of the model-based approach in robotics, model-based algorithms have two fundamental challenges: model fidelity and tractability.
The performance of model-based approaches relies heavily on the accuracy of the model. However, identifying accurate models is often laborious or intractable because of their large search space and nonlinearity. Other than the model accuracy, models also need to be computationally feasible for real-time control applications. High-fidelity but highly complex models are often difficult to integrate into real-time safety-critical driving systems.

To tackle these challenges, conventional approaches, including the Prediction Error Method, have been used to identify model parameters\cite{tangirala2018principles}.
However, those methods often require the model structure to be linear or in specific mathematical forms, which might not be feasible to describe the nonlinear high-speed autonomous driving system.
Gradient-based optimization (GBO)\cite{chong2013introduction} methods are often employed for parameter identification by updating parameters with a function's gradient vector. However, GBO methods can be sensitive to the initial guess and learning rate of parameter updates. They also necessitate a differentiable model and objective function.
On the other hand, another traditional approach, particle swarm optimization (PSO)\cite{erdougmucs2016nonlinear} is a stochastic optimization that adjusts candidate solutions iteratively without requiring differentiability. Nevertheless, it can be highly affected by the choice of its parameters, such as the number of particles and acceleration coefficients. Poorly chosen parameter values can lead to slow or premature convergence to suboptimal solutions.
Recent studies have utilized neural networks (NNs) to model nonlinear dynamics, such as using a simple NN to replace a single-track model \cite{spielberg2019neural} or an NN-based model approximator to identify the vehicle dynamics model in an end-to-end learning fashion \cite{hermansdorfer2021end}. However, integrating NNs with non-learning model-based reliable methods in real-world applications can be difficult. Additionally, ensuring the validity of the NN model in unseen driving scenarios without large-scale field tests is challenging.

In this letter, we propose a model parameter identification method via a hyperparameter optimization scheme (MI-HPO) to efficiently optimize the configuration of model parameters for high-speed autonomous racing systems.
Hyperparameter optimization (HPO) techniques are used to select the optimal hyperparameter configuration for a learning algorithm in machine learning. As the learning process can take a substantial amount of time, these techniques prioritize a balanced exploration and exploitation strategy to efficiently select hyperparameters \cite{feurer2019hyperparameter}.
Our key idea is to leverage an HPO approach to identify physics-based parametric models in a data-driven manner without any limitation on the form of the model equation.
To this end, we adopt Hyperband\cite{li2017hyperband}, a novel HPO scheme with an explore-exploit strategy, while incorporating mutation and annealing methods to design the MI-HPO algorithm.
Using our method, we successfully estimate the parameters of the integrable parametric dynamics models for a full-scaled autonomous racecar, Dallara AV-21 (Fig. \ref{fig:method_overview}), at the Indy Autonomous Challenge (IAC) \cite{IAC}.
We further investigate the advantages of MI-HPO by comparing its performance with conventional optimization methods. Finally, We validate our method by integrating the identified models into the high-speed autonomous system and conducting extensive field experiments, including over \SI{200}{km/h} driving and obstacle avoidance scenarios in the Indianapolis Motor Speedway (IMS) and Las Vegas Motor Speedway (LVMS).

The rest of the letter is organized as follows. Sec. II presents the MI-HPO algorithm; Sec. III and IV introduce vehicle dynamics models and model-based planning/control for our racing system, respectively; Sec. V contains our main evaluation results; and finally, Sec. VI concludes the letter.

%% file: sections/3.1_model_learning.tex
\section{Model Parameter Identification}
\label{sec:model_identification}
In this section, we formulate the model parameter identification problem combined with an HPO approach.
First, we regard a parameter configuration $p \in \mathbb{R}^{n_p}$ with $n_p$ model parameters as a set of hyperparameters of a nonlinear dynamics model $f$. Then, we identify the parameter configuration by evaluating the following objective function inspired by the standard supervised learning problem:
\begin{equation}
    \label{eq:objective}
    \mathcal{L} = \frac{1}{\lvert D \rvert} \sum_{(x,y) \in D} ( y - {f}(x; p) )^2,
\end{equation}
where $x, y$ denote the sampled input and output data of the model $f$ from a given dataset $D$. By minimizing this learning objective, we find an optimized configuration $p^*$ that has the minimum model error with the observed model output $y$.
\begin{algorithm}[b]
\caption{MI-HPO Algorithm based on Hyperband}
\textbf{Input: } $R, \eta, D$
\begin{algorithmic}[1]
\State $ s_{max} \leftarrow \lfloor \text{log}_{\eta}(R) \rfloor, B = (s_{max} + 1)R $
 
\For{$s \in \{s_{max},s_{max}-1, ..., 0 \}$}
    \State $n = \lceil \frac{B}{R} \frac{\eta^{s}}{(s+1)} \rceil, r = R\eta^{-s}$
    \State $P = \text{get\_model\_param\_config}(n)$ 
    \For{$j \in \{ 0, ..., s \}$}
    \State $ n_j = \lfloor n\eta^{-j} \rfloor, r_j = r\eta^j $
    \State $ L = \{ \text{eval\_with\_mutation}(p,r_j,D) : p \in P \} $
    \State $ P = \text{select\_top\_k\_config}(P, L, \lfloor n_j/\eta \rfloor) $
    \EndFor
\EndFor
\end{algorithmic}
\textbf{Output: } \text{Optimized parameters $p^*$ with the smallest loss.}
\label{algo:MI-HPO}
\end{algorithm}
The model $f$ has no limitation on its form of the equation. Thus, our method can be used for arbitrary parametric models.

We implement MI-HPO incorporating a bandit-based algorithm, Hyperband\cite{li2017hyperband}, as summarized in Algorithm \ref{algo:MI-HPO}.
It is a variation of a random search algorithm with exploitation-exploration features to find the optimal configuration based on an evaluation loss.
The algorithm requires two arguments: $R$, the maximum amount of resource (e.g., the number of evaluation iterations) that can be allocated to a single configuration, and $\eta$, a value that determines the proportion of the discarded configurations.
The two arguments initially set $s_{max}$ and a budget $B$, which derive $s_{max}+1$ combinations of the values $n$ and $r$. They enable various ratios of exploration and exploitation to find the optimal configuration.
The algorithm consists of two loops.
In the outer loop, as $s$ is gradually decreased, the number of samples $n$ is reduced, and conversely, the number of evaluation resources $r$ is increased. This encourages exploration with more samples in the early optimization stages and enhances exploitation by performing more evaluations with fewer samples as the iteration progresses, allowing for efficient optimization.
In the inner loop, the algorithm compares the loss of each sample and allocates more resources to the samples with lower evaluation losses, excluding the ones with higher losses.
It repeats the sampling and exclusion processes until the last configuration remains.

To adjust the original HPO process to the model parameter optimization, we apply an additional Gaussian mutation process \cite{mutation2019genetic} in the evaluation process to explore new neighboring parameters with lower model loss.
Unlike the original HPO algorithm, which only allocates more resources $r_i$ without modifying initial configurations, our approach adds perturbation with a noise random variable $\epsilon \in \mathbb{R}^{n_p}$ at the selected configuration $p$ after the resource allocation as:
\begin{equation}
    \label{eq:gaussian_mutation}
p_{mut} = p + \sigma \odot \epsilon, \quad \epsilon \sim N(0, I),
\end{equation}
where $\odot$ is the element-wise product and $\sigma \in \mathbb{R}^{n_p}$ is the standard deviation of the exploration noise.
To achieve fine parameter updates, we linearly anneal $\sigma$ from its maximum value to its minimum value over the course of the evaluation.
We define the following three functions for MI-HPO:
\begin{figure}[t]
\centering
\includegraphics[width=0.45\textwidth]{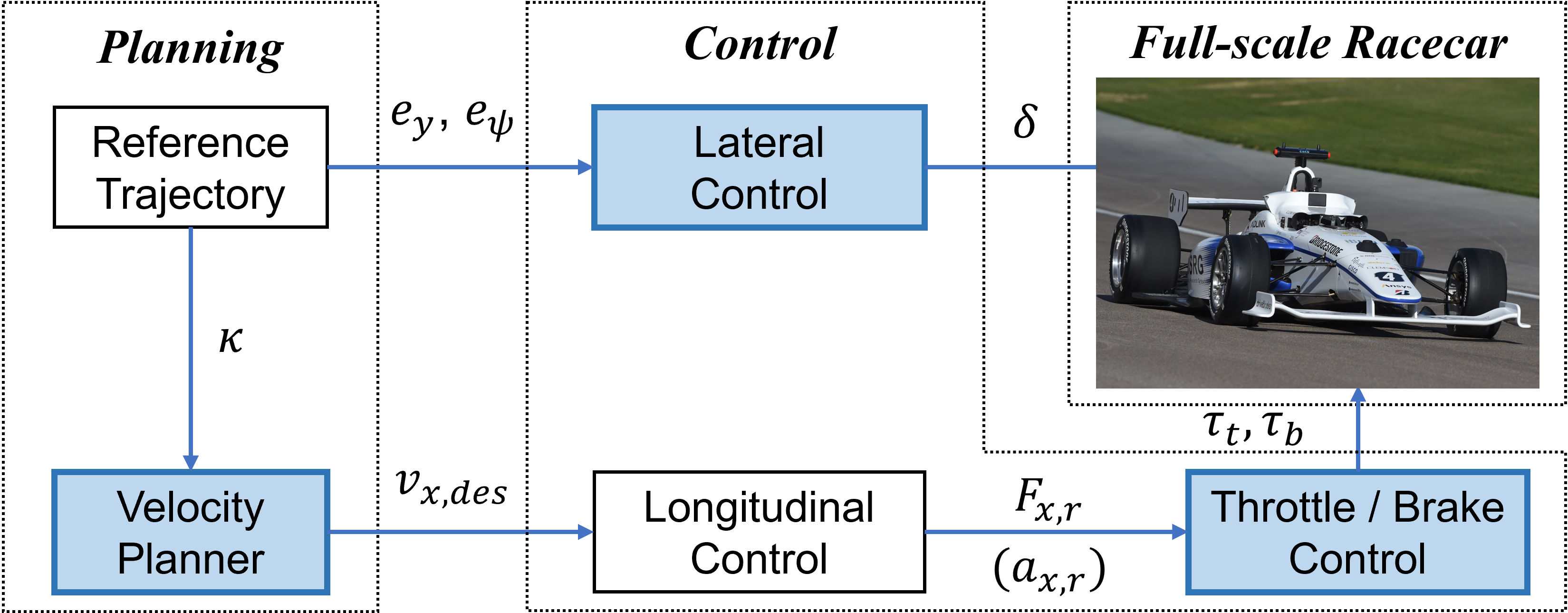}
\caption{
Overview of our autonomous driving system in the AV-21. Our learned model parameters are embedded in the planning and control modules that are covered in this letter (highlighted in blue).
}
\label{fig:method_overview}
\vspace{-1.7em}
\end{figure}
\begin{itemize}
\item \textit{$\text{get\_model\_param\_config}(n)$}\text{:} a function that returns a set of $n$ random parameter configurations from the normal distribution pre-defined over the configuration space.

\item \textit{$\text{eval\_with\_mutation}(p,r_j,D)$}\text{:} a function that receives a parameter configuration $p$, an allocated resource $r_j$, and a dataset $D$ as arguments.
For a total of $r_j$ iterations, the function repeatedly mutates the initial configuration according to Eq. \ref{eq:gaussian_mutation} and evaluates it using Eq. \ref{eq:objective} and $D$.
If a mutated configuration $p_{mut} \in \mathbb{R}^{n_p}$ has a less loss than the initial one, the function replaces $p$ with $p_{mut}$.
It returns the final loss after spending the allocated iterations.
 
\item \textit{$\text{select\_top\_k\_config}(P, L, k)$}\text{:} a function that receives a set of hyperparameter configurations $P$ with their corresponding evaluation losses $L$ and returns the top $k$ high-performing configurations (here, $k = \lfloor n_j/\eta \rfloor$).
\end{itemize}

%% file: sections/3.2_vehicle_dynamics_model.tex
\section{Vehicle Dynamics Model}
\label{sec:vehicle_dynamics_model}
\subsection{Tire Dynamics Model}
The tire model is one of the factors that significantly affect the nonlinearity of driving dynamics. Especially the lateral tire model is crucial to design stable path-tracking control in high-speed driving.
The tire model\cite{bakker1987tyre} can be described as a function of the slip angle $\alpha_i$, slip ratio $\rho_{x,i}$, inclination angle $\theta_i$, tire load $F_{z,i}$, and current velocity $v_{x,i}$, which has a lateral tire force $F^{*}_{y,i}$ of each tire ($i \in \{LF, LR, RF, RR\}$) as,
\begin{equation}
    \label{eq:pacejka_tire_function}
    \begin{aligned}
    F^{*}_{y,i} = f_{tire}(\alpha_i, \rho_{x,i}, \theta_i, F_{z,i}, v_{x,i}).
    \end{aligned}
\end{equation}
Although the model has high fidelity with various dynamics perspectives, it has low suitability for designing the controller of high-speed driving, which requires real-time performance.
Therefore, we first define a tire model with dimension-reduction
that can be applied to model-based control design within an acceptable complexity. We then optimize the model's parameter configuration to represent the overall tire characteristic of a given dataset using our MI-HPO algorithm.
We follow the Pacejka tire model\cite{kabzan2020amz} to define the tire dynamics.
While the prior work neglect vertical and horizontal offsets of the model, we formulate a tire model $F_{y,i} = f_{t,i}(\alpha_i; {p}_{t,i})$ containing the offset parameters $S_{x,i}, S_{y,i}$ to describe the asymmetric tire characteristic determined to maximize cornering performance on an oval track:
\begin{equation}
    \label{eq:tire_model}
    \begin{aligned}
    F_{y,i} &=  D_i \sin(C_i \arctan(B_i (\alpha_i + S_{x,i}))) + S_{y,i},
    \end{aligned}
\end{equation}
where the tire model parameter configuration ${p}_{t,i} = \{B_i, C_i, D_i, S_{x,i}, S_{y,i}\}$ is identified by minimizing the following tire model objective with a given dataset $D_{t,i}$ as
\begin{equation}
    \label{eq:objective_tire}
    \mathcal{L}_{t,i} = \frac{1}{\lvert D_{t,i} \rvert} \sum_{(\alpha_i, F^{*}_{y,i}) \in D_{t,i}} ( F^{*}_{y,i} - f_{t,i}(\alpha; {p}_{t,i}) )^2.
\end{equation}

\subsection{Engine Torque Model}
The powertrain system of our racecar consists of an internal combustion engine, transmission, and wheels. The AV-21 is a rear-wheel-drive vehicle whose traction force $F_{x,r}$ is generated by engine-based driveline dynamics. We model the equation of the longitudinal dynamics\cite{rajamani2011vehicle} as follows:
\begin{equation}
\label{eq:longi_model}
\begin{aligned}
    m a_{x} = F_{x,r} - C_d v^{2}_{x} - C_r,
\end{aligned}
\end{equation}
where $m$ is the vehicle mass, $v_x$ is the longitudinal velocity, and $C_d, C_r$ denote the drag and rolling coefficient, respectively\footnote{The coefficients $C_d, C_r$ were provided by an IAC official.}. Following a prior work \cite{engineSAGE16}, the traction force can be expressed as:
\begin{equation}
\label{eq:traction_force}
\begin{aligned}
    F_{x,r} = m a_{x,r} = \frac{T_{e} \eta_{t} i_{g} i_{0}}{R_w},
\end{aligned}
\end{equation}
where $a_{x,r}$ denotes the traction acceleration, $\eta_{t}$ denotes the efficiency of the transmission, $i_{g}, i_{0}$ denote the transmission ratio of the current gear and final reducer, and $R_w$ denotes the wheel radius. $T_{e} = f_e (w_e, \tau_t)$ is the engine torque map in terms of the engine speed $w_e$ and throttle command $\tau_t$. Due to the high complexity of the engine characteristic, the torque map is expressed as an experimental lookup table based on engine torque curves of specific throttle opening commands.
Although the engine torque model can be obtained by the engine dynamometer testing\cite{killedar2012dynamometer}, it could suffer from modeling errors because the dynamometer testing is done in a static environment. Therefore, we build the engine torque map that integrates the dyno data\footnote{Dyno data refers to the data collected during engine testing on a dynamometer.} with learned torque curves based on our data-driven model identification approach. We express an engine torque curve $T_{e,\tau_{t}} = f_{\tau_{t}}(w_e; {p}_{\tau_{t}})$ of a throttle command $\tau_t$ as a third order polynomial function of the engine speed $w_e$ as:
\begin{equation}
\label{eq:engine_map}
\begin{aligned}
    T_{e,\tau_{t}} = p_{\tau_{t},0} + p_{\tau_{t},1} w_e + p_{\tau_{t},2} w^{2}_e + p_{\tau_{t},3} w^{3}_e,
\end{aligned}
\end{equation}
where $p_{\tau_{t}} = \{p_{\tau_{t},0}, p_{\tau_{t},1}, p_{\tau_{t},2}, p_{\tau_{t},3}\}$ is the torque model parameter configuration.
Using the resultant traction accelerations $a_{x,r}$ that can be derived by Eq. \ref{eq:longi_model} while driving, the engine torque output $T^{*}_{e,\tau_{t}}$ is obtained by Eq. \ref{eq:traction_force}.
Then, $p_{\tau_{t}}$ is learned by minimizing the following engine model objective with a given dataset $D_{\tau_{t}}$:
\begin{equation}
    \label{eq:objective_engine}
    \mathcal{L}_{\tau_{t}} = \frac{1}{\lvert D_{\tau_{t}} \rvert} \sum_{(w_{e}, T^{*}_{e,\tau_{t}}) \in D_{\tau_{t}}} ( T^{*}_{e,\tau_{t}} - f_{\tau_{t}}(w_e; {p}_{\tau_{t}}) )^2.
\end{equation}
To stabilize the learning process, we normalize the engine speed in the range [0,1] with the maximum engine speed.

%% file: sections/3.3_model_based_vehicle_control.tex
\section{Model-based Planning and Control}
\label{sec:model_based_vehicle_control}
In this session, we introduce a model-based planning and control algorithm that uses the learned model parameters. We exploit the learned tire parameters to design a dynamics-aware velocity planning and model-based lateral controller (Fig. \ref{fig:method_overview}). We also integrate the engine dyno data with the learned engine torque model to construct an engine lookup table.
\subsection{Dynamics-aware Velocity Planning}
High-speed cornering during racing causes significant tire load transfer on each wheel due to a lateral acceleration at the roll axis.
Since the tire load governs the maximum performance of the tire, a model-based velocity strategy accounting for the real-time wheel load is necessary to maximize the tire performance without losing tire grip.
We introduce a dynamics-aware velocity planning algorithm that derives the velocity plans with maximum tire performance based on the learned tire dynamics. We first compute the real-time vertical tire load $F_{z,i}$ affected by the lateral load transfer $\Delta W_f$\cite{seward2017race}. The diagram of the load transfer at the roll axis is illustrated in the left of Fig. \ref{fig:method_vehicle_dynamics_control}. The load transfer is computed by the roll couple $C_{roll} = m_s \dot{v}_y h_a$, where $m_s$ is the sprung mass, $\dot{v}_y$ is the lateral acceleration, and $h_a$ is the roll height. As the learned tire model describes the characteristic for the nominal tire load $\bar{F}_{z,i}$, we compute the maximum lateral force of each tire $F^{max}_{y,i}$ in terms of the tire load ratio with peak value of the tire model as:
\begin{equation}
\label{eq:max_lat_tire}
\begin{aligned}
    F^{max}_{y,i} &= \mu \frac{F_{z,i}}{\bar{F}_{z,i}} F^{peak}_{y,i},
\end{aligned}
\end{equation}
where $\mu$ is a tire performance factor to control the confidence and maximum performance of the tire model, $\frac{{F}_{z,i}}{\bar{F}_{z,i}}$ is the tire load ratio.
The maximum lateral acceleration is determined by the following lateral motion dynamics \cite{kabzan2020amz}:
\begin{equation}
\label{eq:lat_accel_limit}
\begin{aligned}
    a_{y,max} = \frac{1}{m} (F^{max}_{y,r} + F^{max}_{y,f} cos(\delta) - m v_x \dot{\psi}),
\end{aligned}
\end{equation}
where $\delta$ is the steering angle and $\dot{\psi}$ is the yaw rate.
Then a desired maximum velocity $v_{x,des}$ is planned according to the curvature $\kappa$ of a reference path from a planning module \cite{lee2022resilient}:
\begin{equation}
\label{eq:velocity_limit}
\begin{aligned}
    v_{x,des} = \sqrt{a_{y,max} / \kappa}.
\end{aligned}
\end{equation}

\subsection{Throttle and Brake Control}
\begin{figure}[t]
\centering
\includegraphics[width=0.48\textwidth]{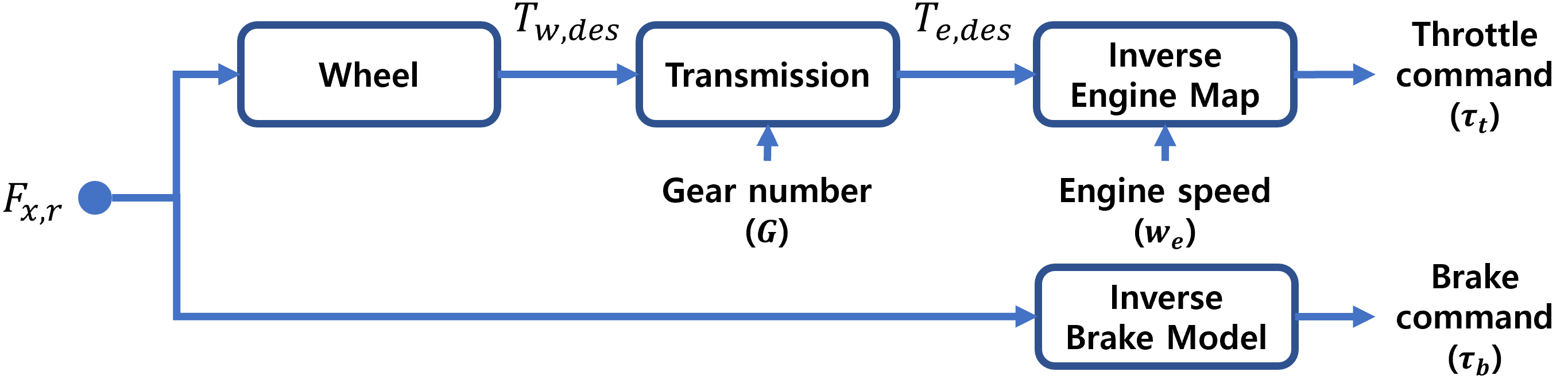}
\caption{
Throttle and brake control system. 
}
\label{fig:method_throttle_brake_control}
\vspace{-1em}
\end{figure}

The planned desired velocity is fed into a feedback control module \cite{doyle2013feedback} to compute the traction force.
However, as shown in Fig. \ref{fig:method_overview}, another low-level controller to transform the traction force to the throttle command is required to control the racecar with nonlinear driveline dynamics. We design the throttle and brake control system following \cite{hedrick1997brake}.
Fig. \ref{fig:method_throttle_brake_control} shows the details of the low-level control system.
We exploit the integrated engine torque map to convert the desired engine torque $T_{e,des}$ to the desired throttle command $\tau_{t,des}$.
As the torque map is built as a lookup table, we search the desired throttle with respect to a given engine speed and desired torque.
The inverse brake model is a module to convert the braking force to the brake pedal command, which is activated if $F_{x,r}$ is negative. The brake model is also attained by our proposed MI-HPO, but details are omitted to conserve space.

\subsection{Model-based Path Tracking Control}
\begin{figure}[t]
\centering
\includegraphics[width=0.46\textwidth]{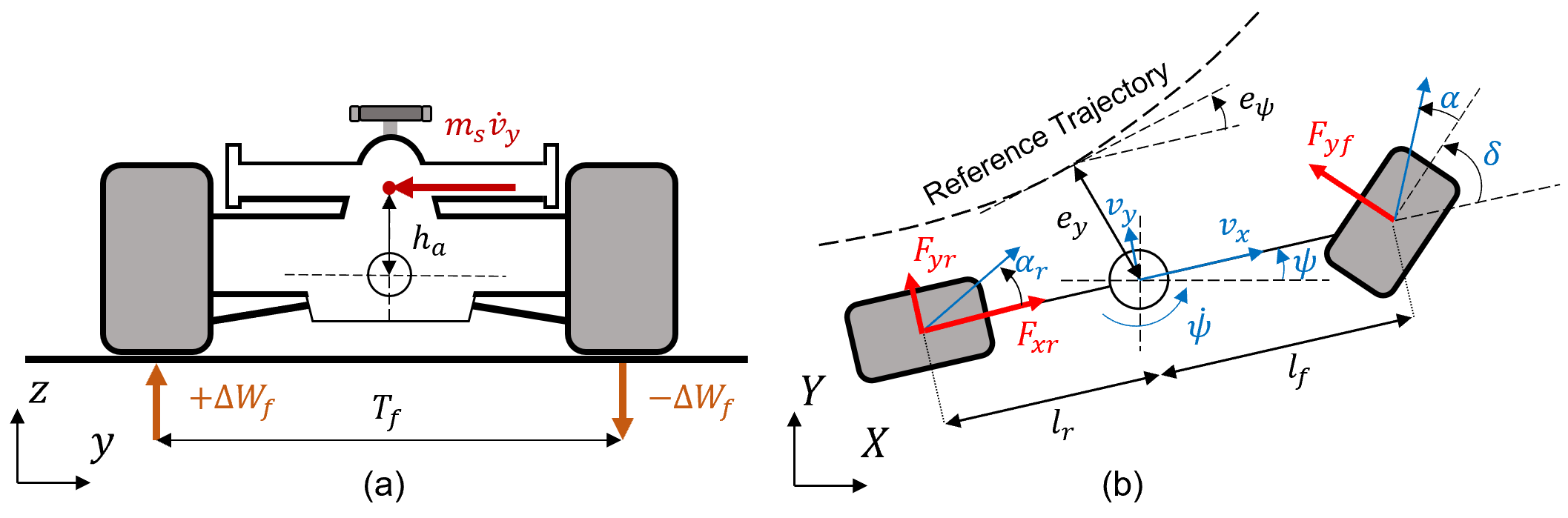}
\caption{
\textbf{Left: } Lateral load transfer generation by the lateral acceleration at the roll axis. 
\textbf{Right: } Overall diagram of the vehicle model.
}
\label{fig:method_vehicle_dynamics_control}
\vspace{-1em}
\end{figure}
We follow the lateral vehicle dynamics of \cite{rajamani2011vehicle} illustrated in the right of Fig. \ref{fig:method_vehicle_dynamics_control}. The lateral model is derived from the objective of tracking a reference trajectory. We implement path tracking control by stabilizing a velocity-dependent chassis model in terms of the error state variables $\xi$ and control $u$.
\begin{equation}
\label{eq:state_control}
\begin{aligned}
    \xi = [e_y, \dot{e}_y, e_{\psi}, \dot{e}_{\psi}]^T, \quad u = \delta,
\end{aligned}
\end{equation}
where $e_y, e_{\psi}$ denote the position and orientation error with respect to a given trajectory. The lateral model contains tire-related model parameters such as the cornering stiffnesses of the front and rear tires $C_{\alpha,f}, C_{\alpha,r}$. Since the lateral dynamics is obtained from the bicycle model, the front and rear tire models are optimized according to Eq. \ref{eq:tire_model}, but the sum of the left and right tire forces is used as the model output. The cornering stiffnesses then can be approximated as follows \cite{rajamani2011vehicle}:
\begin{equation}
\label{eq:cornering_stiffness}
\begin{aligned}
    C_{\alpha f} &\approx B_{f} \times C_{f} \times D_{f}, \quad C_{\alpha r} \approx B_{r} \times C_{r} \times D_{r}.
\end{aligned}
\end{equation}

Based on the lateral model, we design the Linear Quadratic Regulator \cite{lewis2012optimal} with the following optimization problem: 
\begin{equation}
\label{eq:LQR}
\begin{aligned}
    \underset{u}{\text{min }} \int_{0}^{\infty} (\xi^T Q \xi + u^T R u ) dt,
\end{aligned}
\end{equation}
where $Q, R$ denote gain matrices for LQR.
For the real-time control performance, we compute the state feedback optimal LQR gains over piecewise velocity intervals offline\cite{spisak2022robust}.

%% file: sections/4.experiment.tex
\section{Evaluation}
\label{sec:evaluation}
\begin{figure*}[t]
\centering
\includegraphics[width=0.95\textwidth]{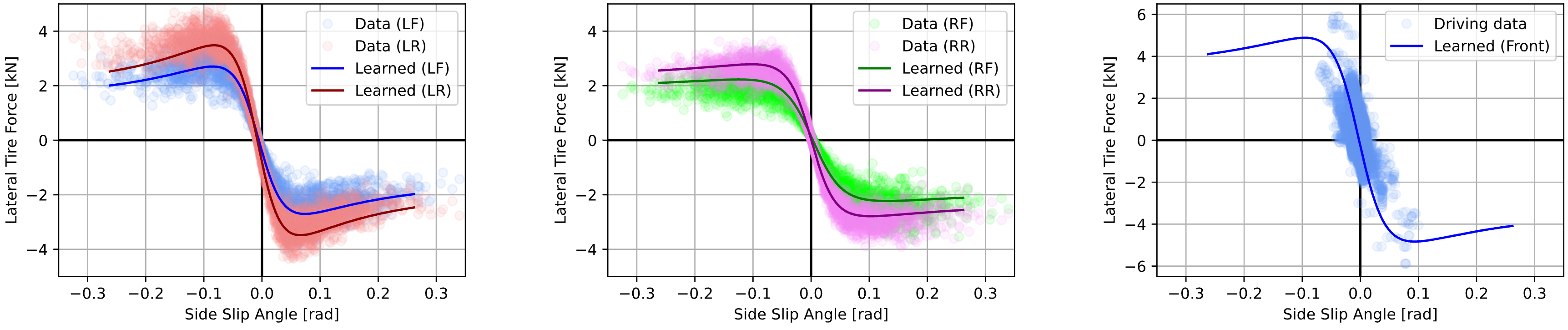}
\caption{
Learned tire models with the provided tire data (\textbf{Left:} left-front and right-front, \textbf{Middle:} left-rear and right-rear). \textbf{Right:} 
Learned front tire dynamics of the single-track bicycle model and the distribution of the collected data on the track.
}
\label{fig:result_tire_model}
\vspace{-1.5em}
\end{figure*}
\begin{figure}[t]
\centering
\includegraphics[width=0.43\textwidth]{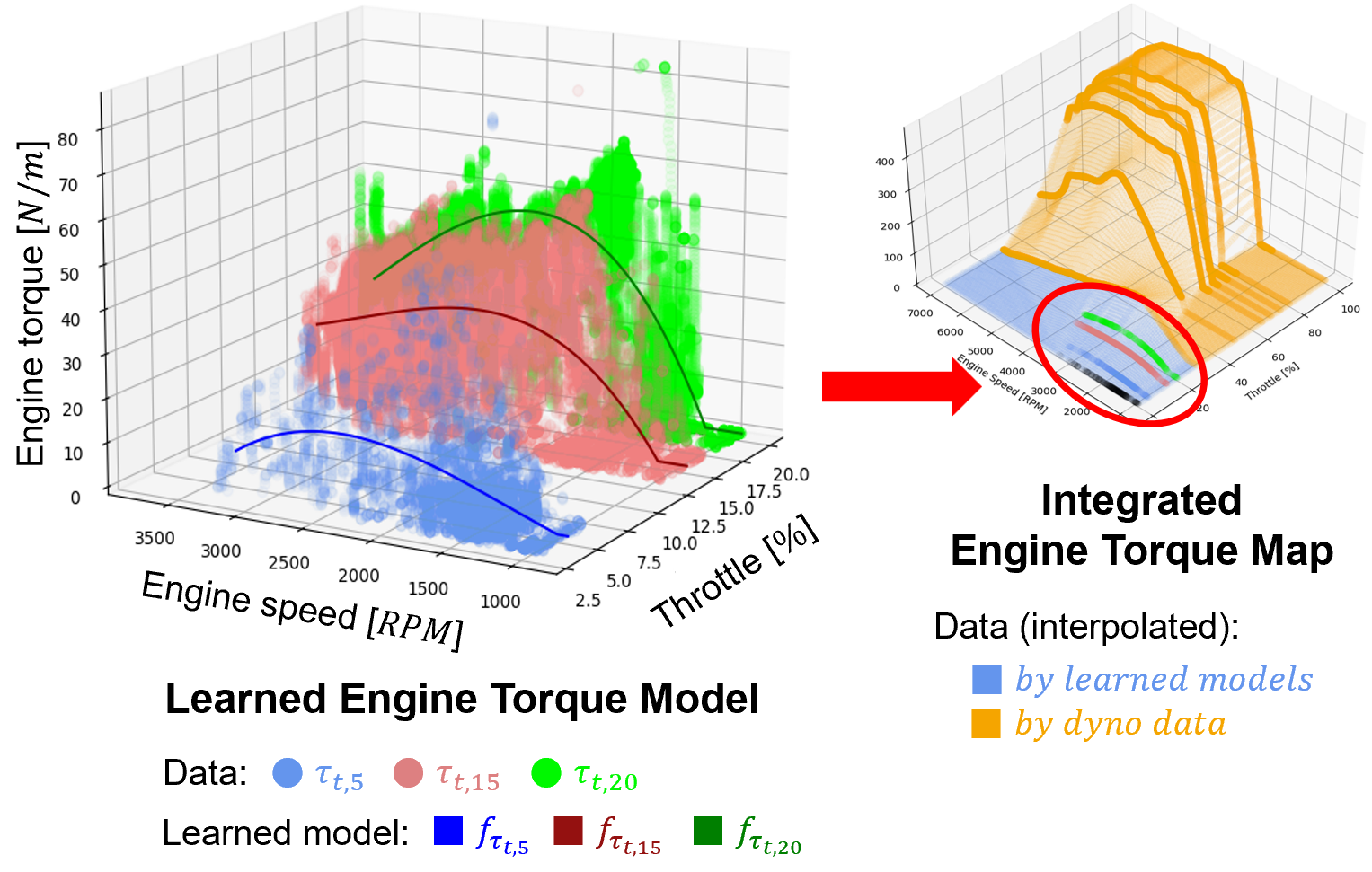}
\caption{
Learned engine torque curves and the integrated engine map.
}
\label{fig:result_engine_brake}
\vspace{-1.5em}
\end{figure}

\subsection{Analysis for Model Parameter Identification}
\subsubsection{Tire Dynamics Model}
Fig. \ref{fig:result_tire_model} illustrates the learned tire models with datasets provided as the tire property files (*.tir)\footnote{The property files for the four tires were provided by an IAC official.}.
The files contain tire force and moment characteristics with high fidelity\cite{schmeitz2013mf}.
To identify the model, we sampled 3000 data for each tire using the property files in various tire state conditions such as tire load, camber angle, slip angle, and slip ratio.
As depicted in the left and middle of Fig. \ref{fig:result_tire_model}, the learned models show good fitness to the tire characteristic distribution.
We further investigated the tire model with the test drive data collected during track racing.
The right of Fig. \ref{fig:result_tire_model} illustrates the front tire model of the single-track bicycle dynamics learned by the provided tire property data comparing it with the driving data that is not used for learning.
The learned model shows the generalization ability for the overall data distribution represented by blue dots.
However, since we obtained the model by offline optimization and focused on the representativeness of data, the model needs higher accuracy in some edge cases near the peaks of the lateral force.
To handle these cases, we can create an online parameter optimization scheme by parallelizing the HPO process\cite{li2017hyperband} in MI-HPO, which we will implement in future work.

\subsubsection{Engine Torque Model}
Fig. \ref{fig:result_engine_brake} illustrates the learned engine torque curves and integrated engine map. The data for the engine map was provided by engine dynamometer testing. 
For higher reliability, we incorporated our data-driven engine torque models with the dyno data, especially for the throttle pedals 5, 15, and 20\%, where the dynamometer showed insufficient accuracy in torque measurements.
The result shows that the learned torque curves are able to represent the change in the maximum torque according to the throttle commands. Moreover, the learned models also fit the torque curves that change nonlinearly as a function of engine speed.
We integrated the learned models with the provided dyno data and interpolated the torque data to construct an engine lookup table. The blue area on the right of Fig. \ref{fig:result_engine_brake} shows the interpolated region by the learned torque curves.
Our vehicle utilized the learned region in racing scenarios such as pit-in/out, obstacle avoidance, and driving within \SI{100}{km/h}.

\subsubsection{Comparison Study}
To evaluate our method's performance, we compared it with four baseline algorithms based on GBO and PSO. We configured two GBO-based methods, namely GBO-small and GBO-large, with learning rates of \SI{5e-12} and \SI{1e-10}, respectively. Additionally, we constructed PSO-100 and PSO-500 with 100 and 500 particles, respectively, utilizing the default acceleration coefficients defined in \cite{erdougmucs2016nonlinear}. For the MI-HPO algorithm, we set $R=\SI{1e+4}{}$ and $\eta=5$. To evaluate the algorithms' performance, we conducted parameter identification for the left-front tire model using the data shown in Fig. \ref{fig:result_tire_model}. All algorithms had the same initial configurations. The average model error curves over time obtained from the evaluation are presented in Fig. \ref{fig:result_param}. Our method outperformed other methods by being 13 and 494 times faster in achieving an error of 1.0 kN and 0.5 kN, respectively. Furthermore, the MI-HPO method achieved a terminal error of 0.39 kN, resulting in a $24\%$ and $76\%$ better convergence than PSO-500 and GBO-large methods, respectively. These results demonstrate the effectiveness of our algorithm in exploring a large number of configurations initially ($n = \SI{1e+4}{}$) and efficiently exploiting well-performing candidates by discarding those with poor performance. GBO-small, with its stable but small learning rate setting, demonstrated negligible improvement in parameter values. Although GBO-large showed better performance than GBO-small, it yielded an unstable optimization process with convergence to a local minimum. PSO-100 had fast initial parameter updates, but it failed to find more optimal parameters beyond an error of 0.75 kN. On the other hand, PSO-500 found a better solution than PSO-100 with more particles, but required higher computational resources due to the increased number of particles. As a result, it had a significantly slower convergence rate compared to our method.

\subsection{Control Performance in Indy Autonomous Challenge}
Our model-based planning and control algorithms were deployed in the full-scale racecar platform. Moreover, we extensively validated our learned parameter-based algorithms in the real-world tracks, IMS and LVMS.
Our vehicle successfully performed various scenarios, such as obstacle avoidance and high-speed racing over {\SI{200}{km/h} on the tracks.

\subsubsection{Obstacle Avoidance at the IMS}
Fig. \ref{fig:result_IMS_obstacle_only} shows the quantitative results of an obstacle avoidance mission
on the front stretch of the IMS.
In this mission, obstacles are located before the first-corner section, where the velocity plan is critical for avoiding collision while keeping close to the racing line. For the sake of safety, we set the tire performance factor $\mu$ as 0.7.
Our dynamics-aware velocity planner was able to allow the racecar to maximize the velocity while regulating the lateral acceleration within the learned maximum tire performance during the rapid avoidance maneuvers.
The obstacle avoidance was initiated in high-speed driving at \SI{100}{km/h}, and steering commands were computed up to $-10\degree$ to follow the generated collision-free reference trajectory.
The sharp steering command could cause significant lateral acceleration higher than \SI{9.0}{m/s^2}, which might cause critical tire grip loss. 
However, our tire model-based velocity planner was able to plan a safe desired speed that did not excessively increase lateral acceleration, and thereby prevent significant loss of tire grip, by inferring a reasonable maximum lateral acceleration based on the real-time tire loading.

\begin{figure}[t]
\centering
\includegraphics[width=0.48\textwidth]{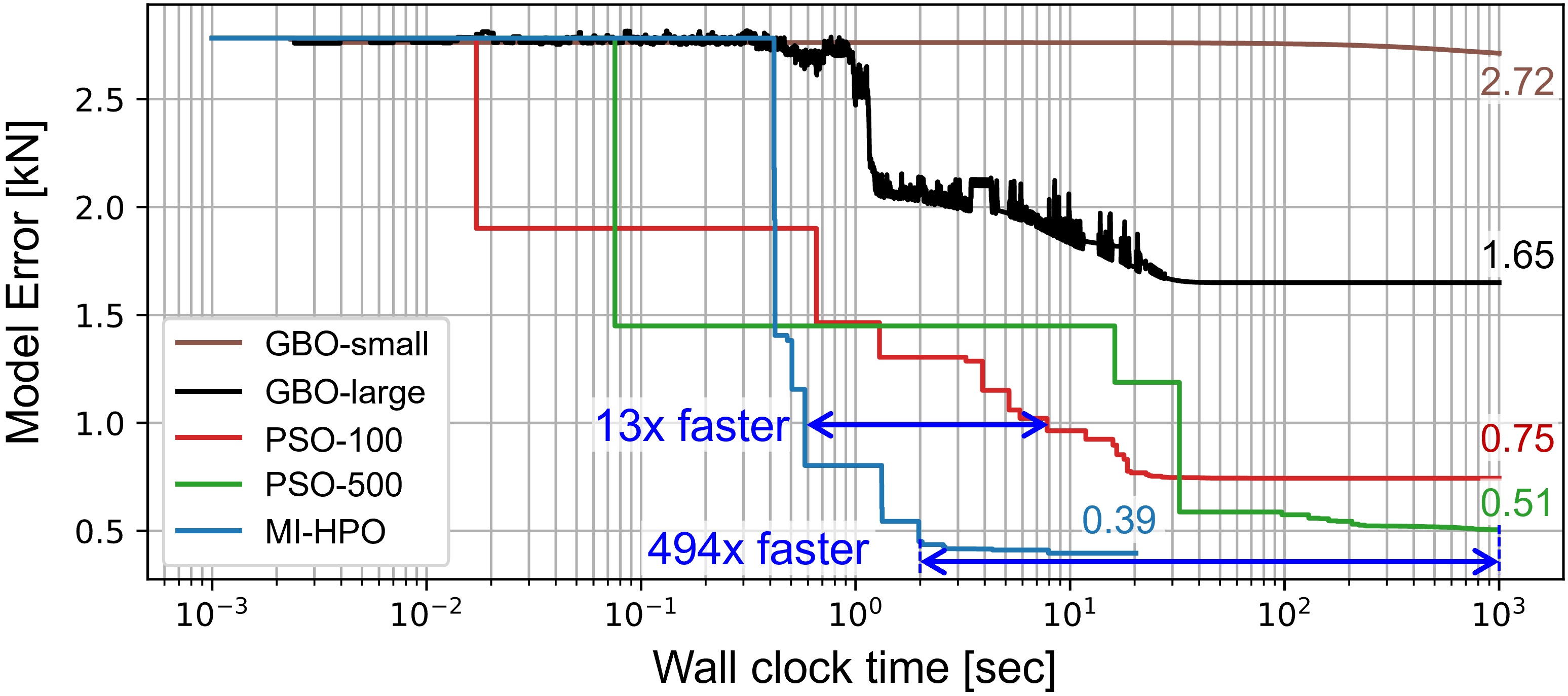}
\caption{
Average model error curves of different evaluation algorithms with the final model error corresponding to each color-coded number.
}
\label{fig:result_param}
\vspace{-1.5em}
\end{figure}

\subsubsection{High-Speed Autonomous Driving in LVMS}
Furthermore, we extensively validated the control performance based on the optimized model parameters at LVMS.
Fig. \ref{fig:result_LVMS_tracking_perform_only} illustrates the quantitative results of the lateral and longitudinal control while our vehicle raced more than nine laps (\SI{23}{km}). Our path-tracking algorithm shows robust control performance leveraging the learned tire parameters. The largest position and orientation errors were \SI{0.6}{m} and $-2.2\degree$, respectively.
In addition, the AV-21 succeeded high-speed autonomous driving at above \SI{144}{km/h} (with a top speed of \SI{217.4}{km/h}), where the dynamic scenario had yet to be visited and adjusted before this track experiment.
These results demonstrate that MI-HPO can optimize and provide appropriate prior dynamics models offline for the design of model-based control before deployment.
However, after the vehicle surpassed \SI{144}{km/h}, our low-level controller computed throttle commands of over 40\%, where the engine range was based solely on the provided dyno data, without our data-driven identification. This could have contributed to the velocity error observed in the range of 151-\SI{187}{km/h} (42-\SI{52}{m/s} in Fig. \ref{fig:result_LVMS_tracking_perform_only}).
Nevertheless, the control system can be improved with an extended data-driven engine map for throttle control at high-speed.
We also point out that our method has the potential to be processed online by parallelizing the HPO process\cite{li2017hyperband}, which enables the method to identify the model in real-time during deployment.

\begin{figure}[t]
\centering
\includegraphics[width=0.48\textwidth]{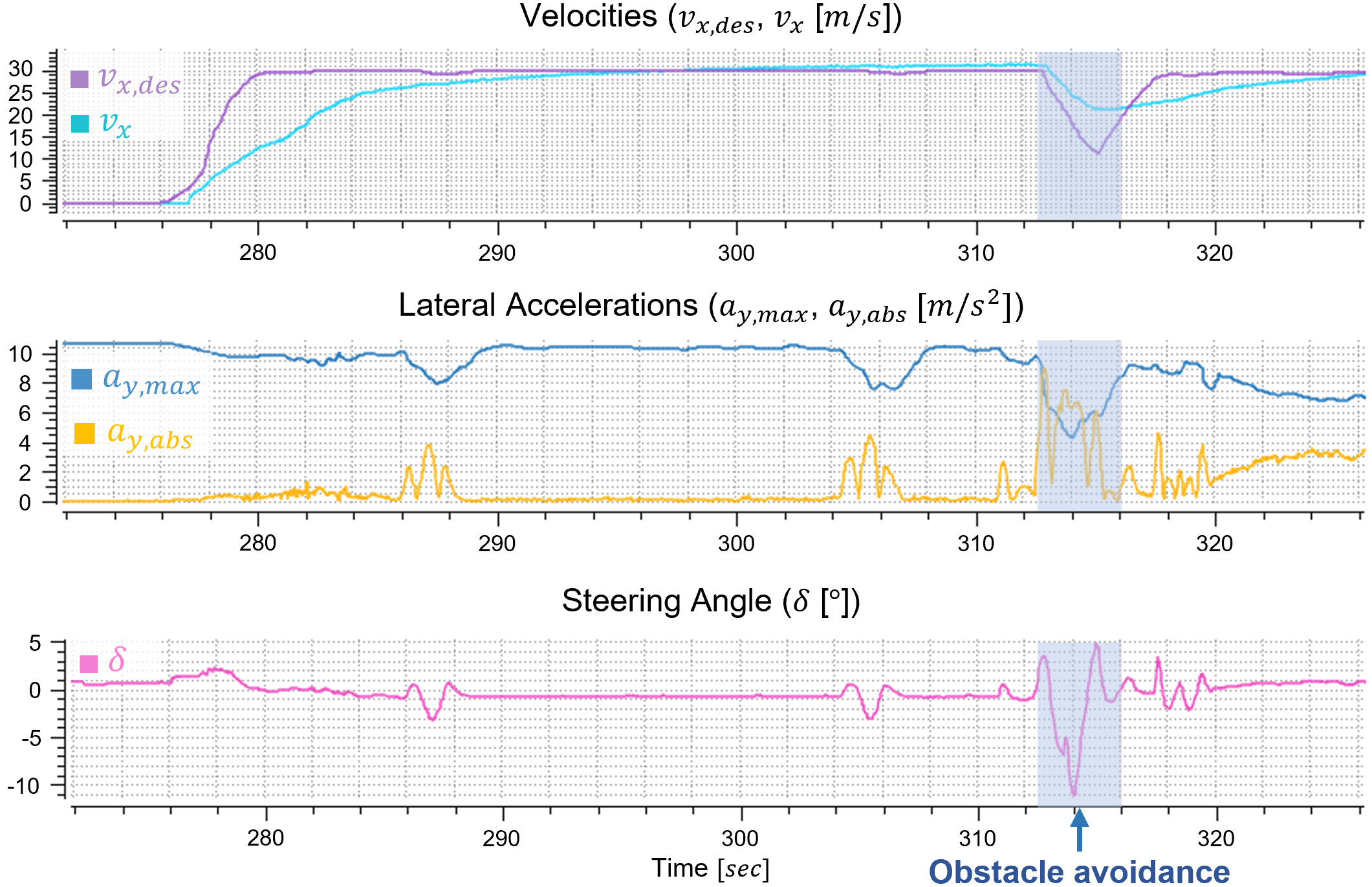}
\caption{
Results of the velocity control, lateral accelerations, and steering angles during the obstacle avoidance mission at the IMS.
}
\label{fig:result_IMS_obstacle_only}
\vspace{-1.3em}
\end{figure}

%% file: sections/5.conclusion.tex
\section{Conclusion}
\label{sec:conclusion}
We present MI-HPO, a method for identifying model parameters by utilizing a hyperparameter optimization scheme.
Our approach showed the ability to optimize the parameters of the dynamics models, such as the tire models and engine torque curves, and exhibited better convergence performance than the traditional optimization schemes. Furthermore, the model-based planning and control system with the learned model parameters demonstrated stable performance in the real-world track environments, IMS and LVMS. In future works, we will implement the online HPO method and integrate it with the offline method of this work to iteratively infer the changing parameters of the vehicle dynamics while on track.